\documentclass[10pt,twocolumn,letterpaper]{article}

\usepackage{wacv}              
\usepackage[accsupp]{axessibility}  

\usepackage{bbm}
\usepackage{multirow}
\usepackage{makecell}
\usepackage{booktabs}
\usepackage{amsfonts}
\newcommand{\figref}[1]{\mbox{Fig.~\ref{#1}}}
\newcommand{\tabref}[1]{\mbox{Tab.~\ref{#1}}}

\definecolor{wacvblue}{rgb}{0.21,0.49,0.74}
\usepackage[pagebackref,breaklinks,colorlinks,allcolors=wacvblue]{hyperref}

\def\wacvPaperID{2828} 
\def\confName{WACV}
\def\confYear{2026}

\title{
KFS-Bench: Comprehensive Evaluation of Key Frame Sampling in Long Video Understanding
}

\author{Zongyao Li \quad Kengo Ishida \quad Satoshi Yamazaki \quad Xiaotong Ji \quad Jianquan Liu\\
Visual Intelligence Research Laboratories, NEC Corporation\\
{\tt\small \{zongyao-li, kengo-ishida, s-yamazaki31, xiaotong-ji, jqliu\}@nec.com}
}

\begin{document}
\maketitle
\begin{abstract}
We propose KFS-Bench, the first benchmark for key frame sampling in long video question answering (QA), featuring multi-scene annotations to enable direct and robust evaluation of sampling strategies. Key frame sampling is crucial for efficient long-form video understanding. In long video QA, selecting informative frames enables multimodal large language models (MLLMs) to improve both accuracy and efficiency. KFS-Bench addresses the limitation of prior works that only indirectly assess frame selection quality via QA accuracy. By providing ground-truth annotations of multiple disjoint scenes required per question, KFS-Bench allows us to directly analyze how different sampling approaches capture essential content across an entire long video. Using KFS-Bench, we conduct a comprehensive study of key frame sampling methods and identify that not only sampling precision but also scene coverage and sampling balance are the key factors influencing QA performance. Regarding all the factors, we design a novel sampling quality metric that correlates with QA accuracy. Furthermore, we develop a novel key frame sampling method that leverages question–video relevance to balance sampling diversity against question–frame similarity, thereby improving coverage of relevant scenes. Our adaptively balanced sampling approach achieves superior performance in both key frame sampling and QA performance. The benchmark is available at \url{https://github.com/NEC-VID/KFS-Bench}.
\end{abstract}    
\section{Introduction}
\label{sec:1}
In recent years, multimodal large language models (MLLMs) have achieved remarkable progress in video understanding~\cite{maaz2023video,lin2023video,lin2024vila,zhang2024video}. Nevertheless, their ability to process long-form videos remains severely constrained by limited input length. Directly applying dense frame sampling to long videos results in prohibitively large token sequences, incurring excessive memory consumption and inference latency. A common workaround is to perform sparse sampling at fixed temporal intervals within an acceptable computational budget. However, such uniform sampling often misses critical information, leading to failures in correctly answering user queries. To address this challenge, two main research directions have emerged. The first focuses on learning additional modules to compress visual tokens under a fixed frame budget~\cite{li2024llama,song2024moviechat,shen2024longvu}, while the second employs key frame sampling to reduce the number of input frames without discarding essential content~\cite{tang2025adaptive,liu2025bolt,yu2025frame,yao2025generative}. Since most state-of-the-art models do not incorporate token-compression modules and training such modules introduces substantial overhead, model-agnostic key frame sampling has become a particularly flexible and practical solution.

\begin{figure}[t]
    \centering
    \includegraphics[width=0.47\textwidth]{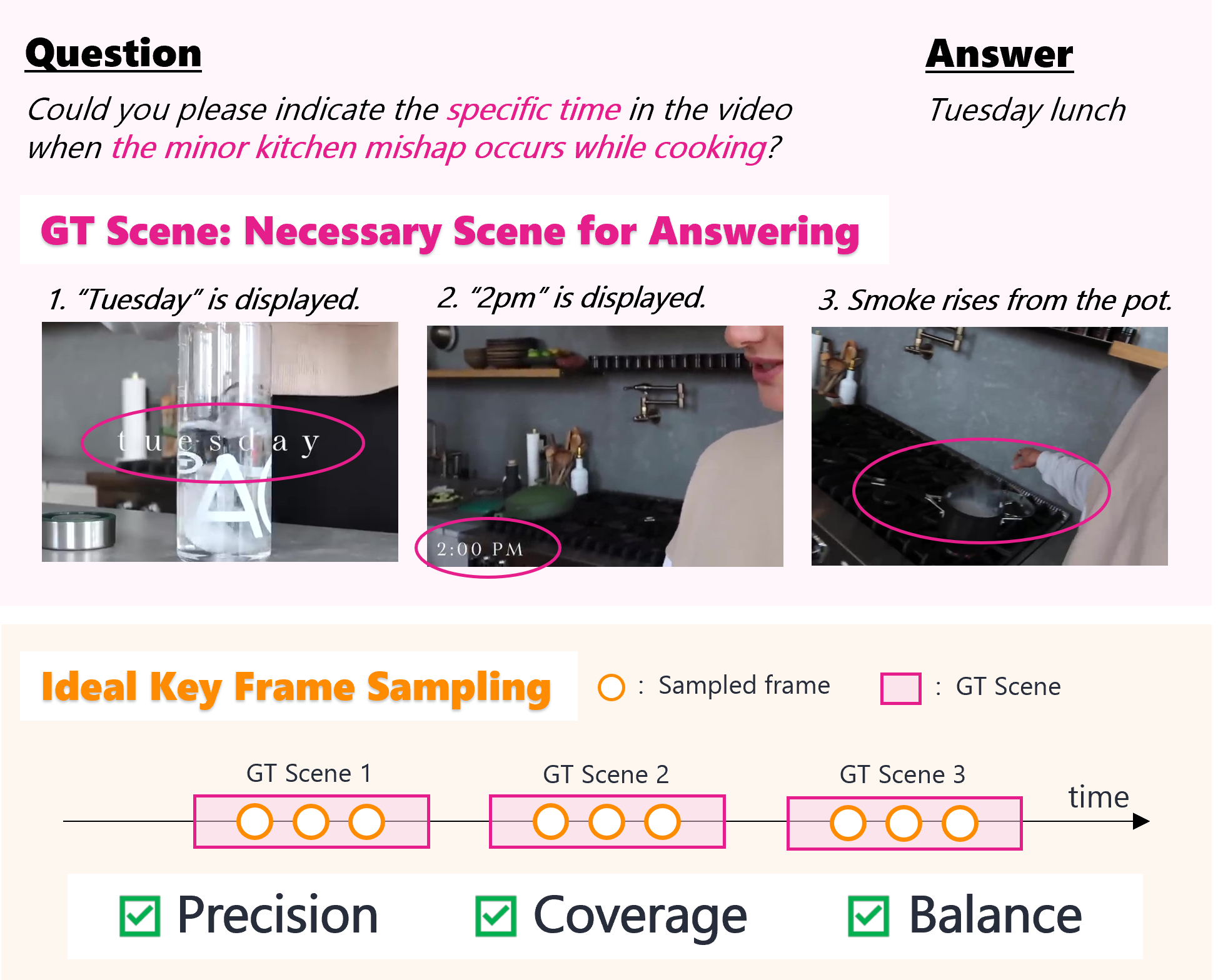}
    \vspace{-0.5em}
    \caption{Ground truth annotation of multiple scenes (GT scenes) and an example of ideal key frame sampling in our KFS-Bench. The ideal key frame sampling, as analyzed in this study,  ensures all extracted key frames fall within GT scenes (Precision), every GT scene is represented (Coverage), and sampling is evenly distributed per scene (Balance). Our analysis shows all the factors play important roles in improving QA accuracy with MLLMs. }
    \label{fig1}
\end{figure}

Existing key frame sampling methods have been primarily developed and evaluated on long video question answering (QA) benchmarks such as LongVideoBench~\cite{wu2024longvideobench} and VideoMME~\cite{fu2025video}. Since these datasets lack explicit annotations of key frames, sampling quality has been assessed only indirectly through QA accuracy. In particular, methods based on question–frame similarity (e.g., using CLIP~\cite{radford2021learning}) determine their sampling strategies and hyperparameters according to QA performance. However, the best QA accuracy does not necessarily correspond to the most effective sampling of key frames, as hallucinations and errors in MLLMs introduce randomness into how sampling affects QA outcomes. This limitation underscores the importance of directly evaluating key frame sampling. While recent benchmarks such as Next-GQA~\cite{xiao2024can} and GroundVQA~\cite{di2024grounded} provide annotations of key segments, they are restricted to short videos and associate each question with only a single segment. In contrast, long video QA often requires reasoning over multiple scenes—actions, events, or situational contexts—that may be distributed across disjoint segments. Therefore, multi-scene annotations are essential for evaluating the general effectiveness and robustness of key frame sampling methods.

To address the above limitations, we introduce \textbf{KFS-Bench}, the first benchmark for key frame sampling in long video QA with multi-scene annotations. Building upon existing benchmarks LongVideoBench and VideoMME, we annotate the question–video pairs with the scenes required to answer each question. As illustrated in \figref{fig1}, for questions that require reasoning over multiple temporally disjoint segments, these segments are labeled as belonging to the same or different scenes depending on the key information they provide for answering the question. With these multi-scene annotations, we are able to conduct a detailed analysis of the factors in key frame sampling that most strongly influence QA performance, and further propose a metric for comprehensive evaluation of sampling quality. In addition, we investigate the correlation between the proposed metric and QA accuracy, thereby demonstrating the effectiveness and reliability of our evaluation framework.

Beyond KFS-Bench, we further propose a novel adaptive similarity-clustering sampling method. Existing methods based on question–frame similarity focus primarily on temporal coverage; however, since different scenes vary in length, broader temporal coverage does not necessarily yield better coverage of relevant scenes. Moreover, these methods typically adopt the same sampling strategy across all questions, overlooking the fact that the degree of relevance between a question and the video content can differ substantially. For example, questions such as “\textit{Which of the following scene sequences is correct?}” contain little direct visual information, rendering question–frame similarity ineffective. To address these issues, we estimate a question–video relevance score from question–frame similarity and use it to adaptively balance the emphasis between similarity-driven sampling and sampling diversity. The latter is achieved by sampling according to an inverse cluster frequency distribution derived from frame clustering, thereby improving coverage of relevant scenes. Experimental results on LongVideoBench and VideoMME demonstrate that our method outperforms existing methods in both sampling quality and QA accuracy.

Our main contributions are summarized as follows:
\begin{itemize}
    \item We introduce \textbf{KFS-Bench}, the first benchmark for long video QA with multi-scene annotations, enabling direct evaluation of key frame sampling.
    \item We conduct a systematic analysis of the factors in key frame sampling that affect QA performance, and based on these insights, propose a comprehensive evaluation metric. The effectiveness of this metric is validated through large-scale experiments showing its correlation with QA accuracy.
    \item We propose a novel key frame sampling method that leverages question–video relevance to balance the emphasis between similarity-driven sampling and sampling diversity.
    \item Experiments on LongVideoBench and VideoMME demonstrate that our method achieves superior performance in both sampling quality and QA accuracy compared to existing methods.
\end{itemize}
\section{Related Work}
\label{sec:2}

\subsection{Benchmarks for Long Video Understanding}
\label{sec:2.1}
LongVideoBench~\cite{wu2024longvideobench} and VideoMME~\cite{fu2025video} are recent benchmarks designed to evaluate MLLMs on long video understanding tasks. LongVideoBench designed multiple-choice QA tasks to test MLLMs' capability in detail retrieval and contextual reasoning. VideoMME features videos from six main genres and 30 subcategories for broad generalization. Both benchmarks show that top models like GPT-4 variants~\cite{achiam2023gpt} and Gemini 1.5 Pro~\cite{team2024gemini} struggle with longer, unedited videos—accuracy drops as video length increases due to limited temporal context. These benchmarks inherently require models to identify important frames or sections within hours of footage, even though the focus is on question answering.

Next-GQA~\cite{xiao2024can}, GroundVQA~\cite{di2024grounded}, and RexTime~\cite{chen2024rextime} are video QA benchmarks that provide key segment annotations. However, these benchmarks are limited to short videos and link each question to only one segment. In contrast, long-duration video QA often requires associating questions with multiple, non-overlapping segments. Multi-hop EgoQA~\cite{chen2025grounded} addresses this issue by offering multi-segment annotations per question for the Ego4D~\cite{grauman2022ego4d} dataset; however, the videos analyzed remain limited to a maximum duration of tens of minutes. Other video understanding benchmarks, including MVBench~\cite{li2024mvbench} for general video understanding, EgoSchema~\cite{mangalam2023egoschema} for egocentric narratives, ActivityNet-QA~\cite{yu2019activitynet} for activity-centric QA, LVBench~\cite{wang2024lvbench} for extremely long videos, VideoEspresso~\cite{han2025videoespresso} for reasoning-oriented video QA with multimodal and CoT annotations, and Narrative Dataset~\cite{wong2022compute,stephen2022narrative} for user-centric narrative summarization, broaden the evaluation landscape but focus on different aspects of video comprehension. Nevertheless, current benchmarks have not comprehensively examined the impact of key frame selection on long video QA accuracy. Indeed, the critical factors of key frame sampling to improve long video QA accuracy remain unclear.

To address these limitations, our KFS-Bench introduces multi-scene annotations for each question in long-form videos of up to approximately one hour, utilizing LongVideoBench and VideoMME as the sources. Our analysis on KFS-Bench identifies key factors in effective key frame sampling that contribute to improved accuracy in long video QA. In consideration of these factors, we have developed a novel sampling quality metric that correlates with QA accuracy. This metric enables researchers to objectively assess and compare different frame sampling methods, thereby advancing MLLMs for long-form video understanding.

\subsection{Key Frame Sampling  Approaches for Long Video QA}
\label{sec:2.2}
To tackle the above challenge of long video inputs, recent works have proposed sophisticated key frame sampling approaches as part of their long video understanding pipelines~\cite{tang2025adaptive,liu2025bolt, yao2025generative, yu2025frame}. Instead of naive uniform sampling of frames, these methods aim to pick the most informative frames related to the query contents.

For instance, BOLT~\cite{liu2025bolt} proposes Inverse Transform Sampling (ITS), a probabilistic technique that generates samples based on a query-guided probability distribution. Adaptive Keyframe Sampling (AKS)~\cite{tang2025adaptive} improves key frame selection by balancing the trade-off between relevance and coverage. Both methods are training-free approaches, and use heuristic scoring based on the similarity between questions and frames to prioritize frames that are most likely to be relevant.
On the other hand, GenS~\cite{yao2025generative} trains VideoLLM to identify key frames by generating relevance scores assigned to each frame. Frame-Voyager~\cite{yu2025frame} learns informative set of frames by leveraging combinations of ranked frames as training data. These methods estimate relevant frames through optimization and learning. In parallel, person-centered caption-based scene retrieval~\cite{ishida2025person} demonstrates that structured, entity-centric representations improve long-video understanding, especially in multi-person and multi-camera settings.

Although the existing approaches demonstrate the advantages of key frame sampling for long video QA, there has not been explicit evaluation of key frame sampling accuracy. In addition, appropriate ground truth and evaluation metrics tailored to key frame sampling for improving long video QA are lacking.
\section{KFS-Bench Dataset}
\label{sec:3}

\subsection{Data Composition}
\label{sec:3.1}
To enable direct evaluation of key frame sampling in long video QA, we propose KFS-Bench, the first benchmark with multi-scene annotations for long-form videos. We build KFS-Bench by augmenting two representative datasets, LongVideoBench~\cite{wu2024longvideobench} and VideoMME~\cite{fu2025video}, with annotations of the required scenes for answering each question. These datasets are chosen because they span a wide range of video durations—from a few seconds up to one hour—and cover diverse tasks and domains.

Concretely, we annotated all 1,337 video–question pairs from the validation set of LongVideoBench, as well as 900 video–question pairs from VideoMME, sampled from its total of 2,700 pairs. The 900 pairs were selected to be approximately balanced across 11 task types, excluding \textit{Information Synopsis} due to the difficulty and ambiguity of establishing clear annotation standards. After discarding cases deemed unannotatable (e.g., when the relevant segments could not be located in the video) and removing low-quality annotations, the final benchmark consists of 1,291 annotated pairs from LongVideoBench and 888 annotated pairs from VideoMME. The two subsets of the datasets are denoted by \textbf{LongVideoBench}$_{\rm kfs}$ and \textbf{VideoMME}$_{\rm kfs}$, respectively. Together, they form the complete KFS-Bench dataset.

\subsection{Annotation Method and Criteria}
\label{sec:3.2}
Annotators were instructed to label all segments required to answer each question at a 1-second granularity, where the start time was rounded down and the end time rounded up. Each annotated segment was temporally disjoint from others and assigned a unique segment ID. For samples consisting of multiple segments, each segment was assigned a scene ID according to the key information it contributed to answering the question. The annotation followed the principle that if a scene contains multiple segments, the critical information provided by that scene is considered preserved as long as at least one of its segments is sufficiently sampled.

This multi-scene annotation scheme enables direct evaluation of how well sampling strategies capture critical information, and is particularly important for questions that require reasoning across multiple scenes. It thereby provides a more reliable bridge between sampling evaluation and downstream QA performance.

\subsection{Analysis of Key Factors in Frame Sampling Influencing QA Accuracy}
\label{sec:3.3}
To design a comprehensive evaluation metric for key frame sampling that faithfully reflects its impact on QA accuracy, it is necessary to first analyze the sampling factors that influence QA performance. Intuitively, an ideal sampling strategy would (i) include as many key frames as possible, (ii) achieve full coverage of all relevant scenes, and (iii) allocate frames across scenes in proportion to their durations. However, the actual preferences of MLLMs may not align perfectly with these assumptions. To investigate this gap, we leverage ground-truth scene annotations to conduct controlled experiments, analyzing the effect of the following three factors on QA accuracy: (i) proportion of key frames included in the sampled set, (ii) scene coverage, i.e., whether all relevant scenes are represented, and (iii) frame distribution across scenes.

First, to control the first two factors, we design two basic metrics: Key Frame Rate (KFR) and Scene Hit Rate (SHR), defined as follows:
\begin{equation}
{\rm KFR}=\frac{|T \bigcap K|}{|T|},\quad {\rm SHR}=\frac{1}{m}\sum_{i=1}^m \mathbbm{1}[|T_i| \geq 1],
\end{equation}
where $T$ denotes the set of sampled frames, $K$ denotes the ground-truth key frames, $m$ is the number of relevant scenes, and $T_i$ is the set of sampled frames within the $i$-th scene. For the distribution of sampled frames across scenes, we introduce a Dirichlet-based control mechanism. Specifically, we model the sampling proportion vector
\begin{equation}
\mathbf{p}=(p_1,p_2,...,p_m),\quad \sum_{i=1}^m p_i=1,
\end{equation}
as drawn from a Dirichlet distribution with two hyperparameters $C$ and $\beta$:
\begin{equation}
\mathbf{p}\thicksim {\rm Dir}(C \cdot (\frac{l_1^{\beta}}{\sum_{j=1}^m l_j^{\beta}},...,\frac{l_m^{\beta}}{\sum_{j=1}^m l_j^{\beta}})),
\end{equation}
where $l_i$ is the duration of the $i$-th scene. Here, $C$ controls concentration (i.e., how strictly the sampling follows the target distribution), while $\beta$ adjusts the relative weighting of scene durations. By fixing two of these factors and varying the third, we systematically analyze the effect of each factor on QA accuracy.

\begin{table}[t]
    \caption{QA accuracy on \textbf{VideoMME}$_{\rm kfs}$ with different Key Frame Rates and Scene Hit Rates, where the number of frames per scene is proportional to scene duration. The model used is InternVL3-8B~\cite{zhu2025internvl3}. Subtitles are not used.}
    \label{tab1}
    \setlength{\tabcolsep}{1mm}
    \centering
    \begin{tabular}{c|c|cccccc}
        \toprule
        \multicolumn{2}{c|}{} & \multicolumn{6}{c}{Key Frame Rate} \\
        \cmidrule(lr){3-8}
        \multicolumn{2}{c|}{} & 0\% & 20\% & 40\% & 60\% & 80\% & 100\% \\
        \midrule
        \multirow{6}{*}{\rotatebox[origin=c]{90}{Scene Hit Rate}} & 0\% & 53.8 & - & - & - & - & - \\
        & 20\% & - & 64.6 & 66.7 & 67.3 & 66.8 & 67.3 \\
        & 40\% & - & 65.0 & 66.6 & 67.3 & 67.5 & 68.7 \\
        & 60\% & - & 65.3 & 67.6 & 67.8 & 68.2 & 68.8 \\
        & 80\% & - & 65.5 & 67.8 & 68.9 & 69.9 & 71.3 \\
        & 100\% & - & 66.9 & 68.7 & 70.0 & 71.2 & 73.2 \\
        \bottomrule
    \end{tabular}
\end{table}

\subsubsection*{Effect of Key Frame Rate and Scene Hit Rate}
\tabref{tab1} reports QA accuracy on \textbf{VideoMME}$_{\rm kfs}$ under varying KFR and SHR levels, with the frame distribution set proportional to scene durations. Non-key frames were uniformly sampled from the portions of the video that do not belong to any relevant scene. Note that except for the case where SHR = 0\%, each sample was ensured to include at least one scene. As a result, the actual SHR differs from the values reported in the table, though the overall trends remain consistent. Additional experiments following the same protocol are provided in Appendix C, including results on \textbf{LongVideoBench}$_{\rm kfs}$ and with alternative models.

The results demonstrate that, when SHR is controlled, QA accuracy exhibits a strong positive correlation with KFR. Moreover, the higher the SHR, the stronger this correlation tends to be, and the greater the impact of KFR on accuracy (e.g., at SHR = 20\%, the accuracy gap between KFR = 20\% and KFR = 100\% is 2.7\%, whereas at SHR = 100\% the gap increases to 6.3\%). Conversely, when KFR is controlled, SHR shows a similar relationship with QA accuracy: higher KFR strengthens the correlation between SHR and QA accuracy and amplifies the effect of SHR. These findings support our initial hypothesis: effective sampling should aim to include as many key frames as possible while covering as many scenes as possible. Consistent conclusions are observed across the additional experiments reported in Appendix C.

\begin{table}[t]
    \caption{QA accuracy on \textbf{VideoMME}$_{\rm kfs}$ under different distributions of key frames across scenes controlled by the parameters $C$ and $\beta$ in Eq.~(3). The model used is InternVL3-8B~\cite{zhu2025internvl3}. Subtitles are not used.}
    \label{tab2}
    \setlength{\tabcolsep}{1mm}
    \centering
    \begin{tabular}{c|c|cccccc}
        \toprule
        \multicolumn{2}{c|}{} & \multicolumn{6}{c}{$\beta$} \\
        \cmidrule(lr){3-8}
        \multicolumn{2}{c|}{} & 0 & 0.2 & 0.5 & 1 & 2 & 5 \\
        \midrule
        \multirow{6}{*}{$C$} & 0.05 & 68.2 & 68.4 & 68.6 & 68.0 & 68.4 & 68.6 \\
        & 0.2 & 68.6 & 68.0 & 68.7 & 69.0 & 68.7 & 68.2 \\
        & 1 & 71.1 & 70.9 & 69.4 & 69.8 & 70.3 & 69.5 \\
        & 5 & 72.2 & 72.3 & 72.4 & 71.7 & 70.7 & 69.6 \\
        & 20 & 73.1 & 72.7 & 72.7 & 72.4 & 71.2 & 70.7 \\
        \bottomrule
    \end{tabular}
\end{table}

\subsubsection*{Effect of Key Frame Distribution across Scenes}
\tabref{tab2} shows QA accuracy on \textbf{VideoMME}$_{\rm kfs}$ under different distributions of key frames. Both KFR and SHR are fixed at 100\%, while the parameters $C$ and $\beta$ in Eq.~(3) are varied to control the distribution of key frames across scenes. Given the distribution sampled from the Dirichlet prior and the total number of sampled frames, the number of frames assigned to each scene is determined (ensuring at least one frame per scene), after which frames are uniformly sampled within each scene.

From the results in \tabref{tab2}, we first observe that when $\beta$ is fixed, QA accuracy shows a strong positive correlation with $C$. In particular, accuracy varies significantly when $C \geq 0.2$. This indicates that the more evenly frames are distributed, or the more closely the distribution follows scene durations, the higher the QA accuracy. When $C$ is small (e.g., 0.2 or 0.05), sampling produces many extreme cases---such as long-duration scenes receiving very few frames---leading to a notable drop in accuracy.

When $C$ is fixed, we find that for larger values of $C$ (e.g., 5 or 10), QA accuracy drops significantly once $\beta > 1$, while remaining largely unaffected in the range $\beta \in [0,1]$. This suggests that the ideal distribution for MLLMs is not necessarily identical to either the distribution proportional to scene duration or the uniform distribution. Instead, it lies between the two, where any distribution interpolated by $\beta$ within $[0,1]$ approximates an effective sampling strategy. In contrast, when $C$ is small, the prevalence of extreme cases renders accuracy almost independent of $\beta$. These findings are further validated by additional experiments conducted on \textbf{LongVideoBench}$_{\rm kfs}$ and with other models following the same protocol as \tabref{tab2}. The results are provided in Appendix C. These observations refine our initial hypothesis of an “ideal” key frame distribution and enable the design of a more effective metric for evaluating sampling quality.

\subsection{Unified Keyframe Sampling Score}
\label{sec:3.4}
To provide a single comprehensive metric for evaluating key frame sampling that also reflects its impact on QA accuracy, we integrate the three influential factors identified in \ref{sec:3.3} into a unified score, termed the Unified Keyframe Sampling Score (UKSS).

Among these factors, KFR can be directly incorporated into UKSS. However, SHR fails to distinguish extreme cases (e.g., when a long scene receives few frames) from effective coverage. To address this limitation, we replace SHR with Balanced Scene Recall (BSR), as the following equation:
\begin{equation}
{\rm BSR}=\frac{1}{m}\sum_{i=1}^m\mathbbm{1}[|T_i| \geq \theta_i],
\end{equation}
\begin{equation}
\theta_i=\max(1,\lfloor(\sum_{j=1}^m|T_j|)\cdot \min(\frac{l_i}{\sum_{j=1}^m l_j},\frac{1}{m})\rfloor),
\end{equation}
which measures the effective coverage of relevant scenes. A scene is considered effectively covered if the proportion of sampled key frames within the scene is at least proportional to its duration relative to the total, or at least equal to uniform allocation—corresponding to the condition in the function $\mathbbm{1}[\cdot]$ of Eq.~(4).

For the distribution of key frames across scenes, inspired by the observations in \ref{sec:3.3}, we design a metric called Balanced Distribution Similarity (BDS), as follows:
\begin{equation}
{\rm BDS}=\max_{0 \leq \beta \leq 1}\cos(\frac{(|T_1|,...,|T_m|)}{\sum_{i=1}^m |T_i|},\frac{(l_1^\beta,...,l_m^\beta)}{\sum_{i=1}^m l_i^\beta}),
\end{equation}
where $\cos(\cdot,\cdot),$ denotes cosine similarity. BDS evaluates the similarity between the actual key frame distribution and the “ideal” distribution. Based on our experimental findings, the ideal distribution lies approximately between the duration-proportional distribution and the uniform distribution, interpolated by a parameter $\beta \in [0,1]$. Accordingly, BDS is defined as the maximum similarity across interpolated distributions, where $\beta$ is sampled on a discrete grid \{0, 0.1, ..., 1.0\}.

KFR, BSR, and BDS are computed for each sample, and their geometric mean is taken as the sample-level score, which penalizes extremely low values of any component metric. Although the sample score is not linearly correlated with the probability of correct answers, as the relationship may vary across models, question types, and difficulty levels, low scores have a disproportionately strong effect in reducing the likelihood of correct QA. At the same time, even very low scores can occasionally yield correct answers (e.g., in \tabref{tab1}, when KFR and SHR are both 0\%, accuracy remains 53.8\%), so overly severe penalties should be avoided. To balance these considerations, we truncate the sample scores with a small constant $\epsilon$ before aggregation. The final UKSS is then defined as the geometric mean of the truncated sample scores across all samples, as follows:
\begin{equation}
{\rm UKSS}=\sqrt[n]{\prod_{i=1}^n \max(\epsilon,\sqrt[3]{{\rm KFR}_i{\rm BSR}_i{\rm BDS}_i}}),
\end{equation}
where $n$ is the number of samples, and $\epsilon$ is empirically set to 0.01.

\begin{table}[t]
    \caption{Spearman’s $\rho$ between UKSS and QA accuracy under different settings, including datasets, MLLMs, maximum numbers of sampled frames, and question–frame similarity models. Each correlation is computed from 400 evaluations (200 with AKS~\cite{tang2025adaptive} and 200 with ITS~\cite{liu2025bolt}) obtained by varying method-specific hyperparameters. $\rho$: correlation coefficient. \textbf{V}$_{\rm kfs}$: \textbf{VideoMME}$_{\rm kfs}$. \textbf{L}$_{\rm kfs}$: \textbf{LongVideoBench}$_{\rm kfs}$. CLIP: CLIP-L/14~\cite{radford2021learning}. IV2: InternVideo2$_{clip}$-L14~\cite{wang2024internvideo2}.}
    \label{tab3}
    \setlength{\tabcolsep}{1mm}
    \centering
    \begin{tabular}{ccccc}
        \toprule
        \multirow{2}{*}{MLLM} & \multirow{2}{*}{Frames} & \multirow{2}{*}{\makecell{Similarity \\ model}} & \multicolumn{2}{c}{$\rho$} \\
        & & & \textbf{V}$_{\rm kfs}$ & \textbf{L}$_{\rm kfs}$ \\
        \midrule
        Qwen2.5-VL-7B & 32 & CLIP & 0.600 & 0.542 \\
        Qwen2.5-VL-7B & 64 & CLIP & 0.534 & 0.681 \\
        Qwen2.5-VL-7B & 64 & IV2 & 0.836 & 0.885 \\
        InternVL3-8B & 64 & CLIP & 0.697 & 0.678 \\
        \bottomrule
    \end{tabular}
\end{table}

\subsection{Validation of UKSS}
\label{sec:3.5}
To verify the effectiveness of UKSS, we conducted extensive experiments to analyze its correlation with QA accuracy. It is important to note that, due to the inherent uncertainty and errors of MLLMs, even an ideal sampling metric cannot achieve a perfect correlation with QA accuracy. Nevertheless, an effective metric should exhibit a strong correlation.

To demonstrate the generality of UKSS, we evaluated it under different settings, including two models (InternVL3-8B~\cite{zhu2025internvl3}, Qwen2.5-VL-7B~\cite{Qwen2.5-VL}), different maximum numbers of sampled frames (32 and 64), and different question–frame similarity models (CLIP-L/14~\cite{radford2021learning}, InternVideo2$_{clip}$-L14~\cite{wang2024internvideo2}). For each configuration (dataset, MLLM, maximum frame budget, similarity model), we tuned the hyperparameters of two existing methods, AKS~\cite{tang2025adaptive} and ITS~\cite{liu2025bolt}, and performed 400 evaluations in total (200 for each method) on the entire dataset. Specifically, the hyperparameters were sampled as follows:
\begin{equation*}
s_{\rm thr} \in \{0,0.03,...,0.99\},\quad L \in \{1,2,...,6\} \quad {\rm for \ AKS}, \\
\end{equation*}
\begin{equation*}
\alpha \in \{0.05,0.10,...,10.0\} \quad {\rm for \ ITS}.
\end{equation*}
\tabref{tab3} reports the Spearman correlation coefficients between UKSS and QA accuracy across different settings. The results show that, although the correlation is affected by the specific configuration, it consistently remains above 0.5 (moderate correlation) and often approaches or exceeds 0.7 (strong correlation). These findings demonstrate that UKSS provides a reliable measure for evaluating the effectiveness of key frame sampling. The visualization of the results in \tabref{tab3} (scatter plots of accuracy versus UKSS) is provided in Appendix B.
\section{Adaptive Similarity–Clustering Sampling}
\label{sec:4}

\subsection{Limitations of Existing Methods}
\label{sec:4.1}
Current key frame sampling methods based on question–frame similarity either (i) select the top-$k$ most relevant frames (e.g., Goldfish~\cite{ataallah2024goldfish}), or (ii) sample highly relevant frames while expanding temporal coverage (e.g., AKS~\cite{tang2025adaptive}, ITS~\cite{liu2025bolt}). Focusing on high-similarity frames improves the KFR, whereas expanding temporal coverage increases the likelihood of covering required scenes. These two objectives are inherently in a trade-off relationship. However, such balancing strategies face two major limitations:
\begin{itemize}
    \item \textbf{Insufficient sampling diversity.} Existing methods focus only on temporal coverage. Yet, disjoint segments may contain visually similar content, leading to significant redundancy in the sampled frames. In other words, merely expanding temporal coverage does not necessarily result in better effective coverage of required scenes (as measured by BSR introduced in \ref{sec:3.4}).
    \item \textbf{Failure under low question–video relevance.} Not all questions are strongly grounded in the video content. For example, the question “\textit{Which of the following scene sequences is correct?}” contains little direct visual information, making question–frame similarity ineffective. In such cases, existing methods often perform even worse than uniform sampling.
\end{itemize}
To overcome these issues, we propose an Adaptive Similarity–Clustering Sampling (ASCS) method that combines clustering-based sampling with similarity-based sampling. To adaptively balance between the two sampling strategies, we introduce a Question–Video Relevance Score (QVRS), thereby improving effective scene coverage while sacrificing as little KFR as possible.

\subsection{Clustering-Based Sampling}
\label{sec:4.2}
In long video QA, segments containing different key information usually exhibit diverse visual content. This motivates us to employ clustering to roughly group frames into scene-like clusters. Since temporally disjoint segments may contain similar content, we disregard temporal order and apply K-means to partition the set of all frames $I$ into $k$ clusters $\{I_1,...,I_k\}$, where $k$ equals the number of frames to be sampled. We use the same frame features as those employed for computing question–frame similarity. If clustering works ideally---meaning that each required scene contains at least one whole cluster---then sampling one frame per cluster ensures that every required scene is covered by at least one frame. This strategy maximizes sampling diversity, reduces redundancy compared to uniform sampling, and thus improves coverage of required scenes.

When the question is largely unrelated to the video content, this clustering-based method typically outperforms similarity–based sampling. However, question–video relevance is not binary. To make clustering-based sampling more compatible with similarity-based methods, we convert the clustering result into a probability distribution. Specifically, each frame is assigned a sampling probability:
\begin{equation}
P(i)=\frac{1}{|I_{c_i}| \cdot k},
\end{equation}
where $c_i$ denotes the cluster index of frame $i$, and $|I_{c_i}|$ is the size of cluster $c_i$. This yields a normalized distribution, which we call the Inverse Cluster Frequency (ICF) distribution. In this distribution, the larger the cluster, the smaller the probability assigned to each frame within it.

Inspired by ITS~\cite{liu2025bolt}, we do not sample directly from the ICF distribution. Instead, we compute its cumulative distribution function (CDF):
\begin{equation}
F_{\rm icf}(i)=\frac{\sum_{j=1}^i P(j)}{\sum_{j=1}^{|I|} P(j)},
\end{equation}
and then perform uniform sampling on the inverse transform of the CDF:
\begin{equation}
t'_i={\rm argmin}_t\{F_{\rm icf}(t) \geq i/|I|\},\quad i \in \{1,...,|I|\},
\end{equation}
where $t'_i$ is the $t$-th sampled frame. This approximates sampling one frame from each cluster.

\subsection{Question-Video Relevance Score}
\label{sec:4.3}
Intuitively, if a question targets specific segments within a video, the distribution of question–frame similarities tends to be spiky and sparse. In contrast, if the question contains little visual grounding or requires global understanding, the similarity distribution is relatively flat. Motivated by this observation, we propose the question–video relevance score, a metric derived from the distribution of question–frame similarities.

We first normalize the raw similarity scores $s$ using Median Absolute Deviation (MAD) normalization:
\begin{equation}
z_i=\frac{s_i-{\rm median}(s)}{{\rm MAD}(s)}.
\end{equation}
Compared to z-score normalization, MAD normalization is more robust to outliers. The normalized scores are then transformed into a probability distribution $Q(i)$ through a softmax with temperature parameter $\tau$:
\begin{equation}
Q(i)=\frac{\exp(z_i/\tau)}{\sum_j \exp(z_j/\tau)}.
\end{equation}
Based on $Q(i)$, we compute three indicators that measure its concentration:
\begin{itemize}
    \item \textbf{Temporal bin entropy $H_{\rm time}$:} the entropy of the distribution when $Q(i)$ is aggregated into $k$ uniformly divided temporal bins.
    \begin{equation}
    H_{\rm time}=-\sum_{b=1}^k(\sum_{i \in {\rm bin}_b}Q(i)) \cdot \log(\sum_{i \in {\rm bin}_b}Q(i)).
    \end{equation}
    \item \textbf{Mass-based bin entropy $H_{\rm mass}$:} the entropy of temporal span lengths when the cumulative distribution of $Q(i)$ is divided into $k$ equal-mass bins.
    \begin{equation}
    H_{\rm mass}=-\sum_{b=1}^k(\frac{\Delta t_b}{|I|-1}) \cdot \log(\frac{\Delta t_b}{|I|-1}),
    \end{equation}
    where\[
    \left\{
    \begin{aligned}
    &C(i) = \sum_{j=1}^i Q(j), \\
    &u_0=1,\ u_b=\min\{i|C(i) \geq b/k\}\ (b=1,...,k), \\
    &\Delta t_b=u_b-u_{b-1}.
    \end{aligned}
    \right.
    \]
    \item \textbf{Shortest coverage window $L_{\rm cov}$:} the minimal temporal window length required to cover a fixed probability mass on $Q(i)$.
    \begin{equation}
    L_{\rm cov}=\min\{{\rm length}(W)|\sum_{i \in W}Q(i) \geq \gamma\}.
    \end{equation}
\end{itemize}
The geometric mean of these three indicators is used as the final score, controlled by two hyperparameters $\tau$ and $\gamma$:
\begin{equation}
{\rm QVRS}=\sqrt[3]{(1-\frac{H_{\rm time}}{\log k}) \cdot (1-\frac{H_{\rm mass}}{\log k}) \cdot (1-\frac{L_{\rm cov}}{|I|})}.
\end{equation}
QVRS thus measures the informativeness of the question–frame similarity distribution, and can be used to balance the relative importance of similarity-based and clustering-based sampling.

\subsection{Similarity-Clustering Balanced Sampling}
\label{sec:4.4}
For similarity-based sampling, we adopt the existing ITS~\cite{liu2025bolt} method. ITS first normalizes the question–frame similarity scores $s$:
\begin{equation}
s_i'=(\frac{s_i-\min(s)}{\max(s)-\min(s)})^\alpha,
\end{equation}
then computes the CDF from the normalized scores:
\begin{equation}
F_{\rm sim}(i)=\frac{\sum_{j=1}^i s_j'}{\sum_{j=1}^{|I|} s_j'}.
\end{equation}
Finally, frames are uniformly sampled on the inverse transform of the CDF, as in Eq.~(10), but with $F_{\rm icf}$ replaced by $F_{\rm sim}$. In ITS, the hyperparameter $\alpha$ controls the balance between temporal coverage and question–frame similarity: larger $\alpha$ biases sampling toward high-similarity frames, whereas smaller $\alpha$ biases it toward uniform sampling.

Since both our clustering-based sampling and similarity-based sampling adopt inverse transform sampling, we combine them through a weighted interpolation of their CDFs. Specifically, we use the QVRS score as the weight to form a balanced CDF:
\begin{equation}
F_{\rm bal}=(1-{\rm QVRS}) \cdot F_{\rm icf} + {\rm QVRS} \cdot F_{\rm sim}.
\end{equation}
By tuning the parameter $\tau$ in QVRS computation, we can globally adjust the balance between the two strategies: a larger $\tau$ yields a higher QVRS, leading the sampling to favor similarity-based selection. Finally, in Eq.~(10), $F_{\rm icf}$ is replaced with $F_{\rm bal}$ for inverse transform sampling.
\section{Experiments}
\label{sec:5}

\subsection{Datasets}
\label{sec:5.1}
We conducted experiments on two long video QA datasets: LongVideoBench~\cite{shen2024longvu} and VideoMME~\cite{fu2025video}. The original datasets contain 1,337 and 2,700 samples, respectively. Our annotated versions, \textbf{LongVideoBench}$_{\rm kfs}$ and \textbf{VideoMME}$_{\rm kfs}$, consist of 1,291 and 888 samples, respectively. In \textbf{LongVideoBench}$_{\rm kfs}$, the numbers of single-scene and multi-scene samples are 961 and 330, respectively, with multi-scene samples containing an average of 2.5 scenes. In \textbf{VideoMME}$_{\rm kfs}$, the numbers are 614 and 274, respectively, and multi-scene samples contain an average of 3.4 scenes.

\subsection{Implementation Details}
\label{sec:5.2}
All videos were uniformly sampled at 1 fps. Subtitles were not used in all experiments. Question–frame similarities were computed using CLIP-L/14. The details of hyperparameter settings and further analyses are provided in Appendix A.

\begin{table}[t]
    \caption{Experimental results on VideoMME and LongVideoBench (LVB). 
QA accuracy (acc.) is reported on the full set, while UKSS is evaluated on the subsets of KFS-Bench.}
    \label{tab4}
    \setlength{\tabcolsep}{1mm}
    \centering
    \begin{tabular}{ccccccc}
        \toprule
        \multirow{2}{*}{MLLM} & \multirow{2}{*}{Frames} & \multirow{2}{*}{\makecell{Sampling \\ Method}} & \multicolumn{2}{c}{\textbf{VideoMME}} & \multicolumn{2}{c}{\textbf{LVB}} \\
        \cmidrule(lr){4-5}
        \cmidrule(lr){6-7}
        & & & Acc. & UKSS & Acc. & UKSS \\
        \midrule
        \multirow{5}{*}{\makecell{Qwen2.5-\\VL-7B}} & \multirow{5}{*}{32} & Uniform & 60.5 & 0.202 & 57.2 & 0.099 \\
        & & K-means & 60.9 & 0.206 & 59.8 & 0.110 \\
        & & AKS & 62.4 & 0.248 & 61.5 & 0.188 \\
        & & ITS & 62.0 & 0.258 & 61.9 & \textbf{0.202} \\
        & & ASCS & \textbf{63.1} & \textbf{0.265} & \textbf{63.6} & 0.199 \\
        \midrule
        \multirow{5}{*}{\makecell{Qwen2.5-\\VL-7B}} & \multirow{5}{*}{64} & Uniform & 64.0 & 0.276 & 59.5 & 0.143 \\
        & & K-means & 64.7 & 0.282 & 59.5 & 0.157 \\
        & & AKS & 64.5 & 0.291 & 63.2 & 0.235 \\
        & & ITS & 65.2 & 0.313 & 63.4 & 0.256 \\
        & & ASCS & \textbf{66.0} & \textbf{0.318} & \textbf{64.1} & \textbf{0.269} \\
        \midrule
        \multirow{5}{*}{\makecell{Qwen2.5-\\VL-32B}} & \multirow{5}{*}{64} & Uniform & 66.6 & 0.276 & 60.2 & 0.143 \\
        & & K-means & 67.9 & 0.282 & 61.7 & 0.157 \\
        & & AKS & 66.9 & 0.291 & 63.5 & 0.235 \\
        & & ITS & 67.6 & \textbf{0.313} & 63.6 & 0.256 \\
        & & ASCS & \textbf{68.4} & \textbf{0.313} & \textbf{65.5} & \textbf{0.269} \\
        \midrule
        \multirow{5}{*}{\makecell{Intern\\VL3-8B}} & \multirow{5}{*}{64} & Uniform & 65.4 & 0.276 & 60.8 & 0.143 \\
        & & K-means & 66.6 & 0.282 & 62.7 & 0.157 \\
        & & AKS & 66.9 & 0.291 & 64.9 & 0.235 \\
        & & ITS & 67.1 & \textbf{0.313} & 65.1 & 0.256 \\
        & & ASCS & \textbf{67.6} & 0.311 & \textbf{65.4} & \textbf{0.268} \\
        \bottomrule
    \end{tabular}
\end{table}

\subsection{Experimental Results}
\label{sec:5.3}
As shown in \tabref{tab4}, we evaluated five sampling methods under different MLLM settings (Qwen2.5-VL-7B~\cite{Qwen2.5-VL}, Qwen2.5-VL-32B, InternVL3-8B~\cite{zhu2025internvl3}) and maximum frame budgets (32, 64): uniform sampling, K-means (our clustering-based sampling from \ref{sec:4.2}), AKS~\cite{tang2025adaptive}, ITS~\cite{liu2025bolt}, and ASCS (our method). Appendix D reports the results partitioned into single-scene and multi-scene samples.

\subsubsection*{Comparison on QA Accuracy}
The results in \tabref{tab4} show that our method consistently outperforms all other methods across both benchmarks under all settings. This demonstrates the effectiveness of using the proposed question–video relevance score to balance similarity-based and clustering-based sampling. In addition, K-means clustering-based sampling generally performs better than uniform sampling, yielding improvements of up to 1.3\% on VideoMME and 2.6\% on LongVideoBench.

From the results with Qwen2.5-VL-7B, we observe that when the frame budget increases from 32 to 64, the relative improvements of all sampling methods over uniform sampling tend to diminish. This suggests that the benefit of key frame sampling decreases as more frames become available. Moreover, the improvements remain nearly unchanged when moving from Qwen2.5-VL-7B to Qwen2.5-VL-32B, indicating that the effectiveness of key frame sampling is not limited to smaller models.

\subsubsection*{Comparison on UKSS}
In addition to QA accuracy, we also evaluated key frame sampling using the proposed UKSS metric. As shown in \tabref{tab4}, our method achieves the best UKSS values under most settings. On LongVideoBench with Qwen2.5-VL-7B at a frame budget of 32 and on VideoMME with InternVL3-8B at a frame budget of 64, our method yields slightly lower UKSS values than ITS, by 0.03 and 0.02, respectively. This trend does not align with the corresponding improvements in QA accuracy; however, as discussed in \ref{sec:3.5}, the correlation between sampling metrics and QA performance is not complete due to the inherent uncertainty and errors of MLLMs. Furthermore, similar to QA accuracy, K-means clustering-based sampling consistently improves UKSS over uniform sampling. In addition, since reducing the frame budget from 64 to 32 significantly worsens the coverage of required scenes, the UKSS values of all sampling methods drop substantially.
\section{Conclusion}
\label{sec:6}
We presented KFS-Bench, the first benchmark with multi-scene annotations for long video QA, enabling the direct evaluation of key frame sampling. We further proposed UKSS, a comprehensive metric that effectively reflects sampling quality, and introduced ASCS, a key frame sampling method that balances similarity- and clustering-based sampling according to question-video relevance. Experiments on LongVideoBench and VideoMME show that ASCS improves both QA accuracy and UKSS, while UKSS maintains strong correlation with QA performance. We expect KFS-Bench to provide valuable resources for advancing research on long video understanding.
{
    \small
    \bibliographystyle{ieeenat_fullname}
    \bibliography{main}
}

\end{document}


\maketitle
\appendix

\section{Hyperparameter Settings and Analysis}
For \textbf{AKS}, the hyperparameters were set to $s_{\rm thr}=0.2, L=2$ on LongVideoBench and $s_{\rm thr}=0.8, L=5$ on VideoMME. For \textbf{ITS}, the hyperparameter $\alpha$ was set to $7.0$ on LongVideoBench and $2.0$ on VideoMME. In our method \textbf{ASCS}, $\alpha$ was set identically to ITS, and the settings of $\tau$ and $\gamma$ in QVRS computation at the highest QA accuracy are shown in \tabref{supp_tab5}. The results indicate that although the optimal hyperparameters vary across different configurations, they generally fall within certain ranges depending on the dataset. Due to the inherent uncertainty and errors of MLLMs, however, it is difficult for the optimal hyperparameters to remain consistent across models or frame budgets.

In \tabref{supp_tab6}, we analyze the effect of different choices of the hyperparameters $\tau$ and $\gamma$ on QA accuracy and sampling quality (measured by UKSS) when using Qwen2.5-VL-32B. As shown in the table, the parameter setting that achieves the highest QA accuracy ($\tau=0.3$, $\gamma=0.7$) also yields the highest UKSS, and the top three settings in accuracy are included among the top six in UKSS. These results demonstrate a certain degree of correlation between QA accuracy and UKSS, highlighting the value of the proposed metric. In addition, all hyperparameter settings in the table achieve higher accuracy than ITS and AKS, further confirming the stability and effectiveness of our method.

\begin{table}[t]
    \caption{Settings of $\tau$ and $\gamma$ in QVRS computation at the highest QA accuracy.}
    \label{supp_tab5}
    \setlength{\tabcolsep}{0.7mm}
    \centering
    \begin{tabular}{cccccc}
        \toprule
        \multirow{2}{*}{MLLM} & \multirow{2}{*}{Frames} & \multicolumn{2}{c}{VideoMME} & \multicolumn{2}{c}{LongVideoBench}\\
        \cmidrule(lr){3-4}
        \cmidrule(lr){5-6}
        & & $\tau$ & $\gamma$ & $\tau$ & $\gamma$ \\
        \midrule
        Qwen2.5-VL-7B & 32 & 0.4 & 0.7 & 0.2 & 0.6 \\
        Qwen2.5-VL-7B & 64 & 0.6 & 0.6 & 0.2 & 0.7 \\
        Qwen2.5-VL-32B & 64 & 0.5 & 0.6 & 0.3 & 0.7 \\
        InternVL3-8B & 64 & 0.4 & 0.7 & 0.6 & 0.7 \\
        \bottomrule
    \end{tabular}
\end{table}
\begin{table}[t]
    \caption{Analysis of different $\tau$ and $\gamma$ settings on QA accuracy (left) and UKSS (right) when using Qwen2.5-VL-32B.}
    \label{supp_tab6}
    \setlength{\tabcolsep}{1mm}
    \centering
    \begin{tabular}{c|c|cc}
        \toprule
        \multicolumn{2}{c|}{} & \multicolumn{2}{c}{$\gamma$} \\
        \cmidrule(lr){3-4}
        \multicolumn{2}{c|}{} & 0.6 & 0.7 \\
        \midrule
        \multirow{5}{*}{$\tau$} & 0.1 & 64.2 / 0.266 & 64.0 / 0.265 \\
        & 0.2 & 65.4 / 0.266 & 65.2 / \textbf{0.269} \\
        & 0.3 & 65.4 / 0.267 & \textbf{65.5} / \textbf{0.269} \\
        & 0.4 & 64.6 / 0.268 & 65.1 / 0.262 \\
        & 0.5 & 65.1 / 0.264 & 64.2 / 0.262 \\
        \bottomrule
    \end{tabular}
\end{table}

\section{Visualization of Accuracy–UKSS Correlation}
\begin{figure*}[t]
    \centering
    \includegraphics[width=0.8\textwidth]{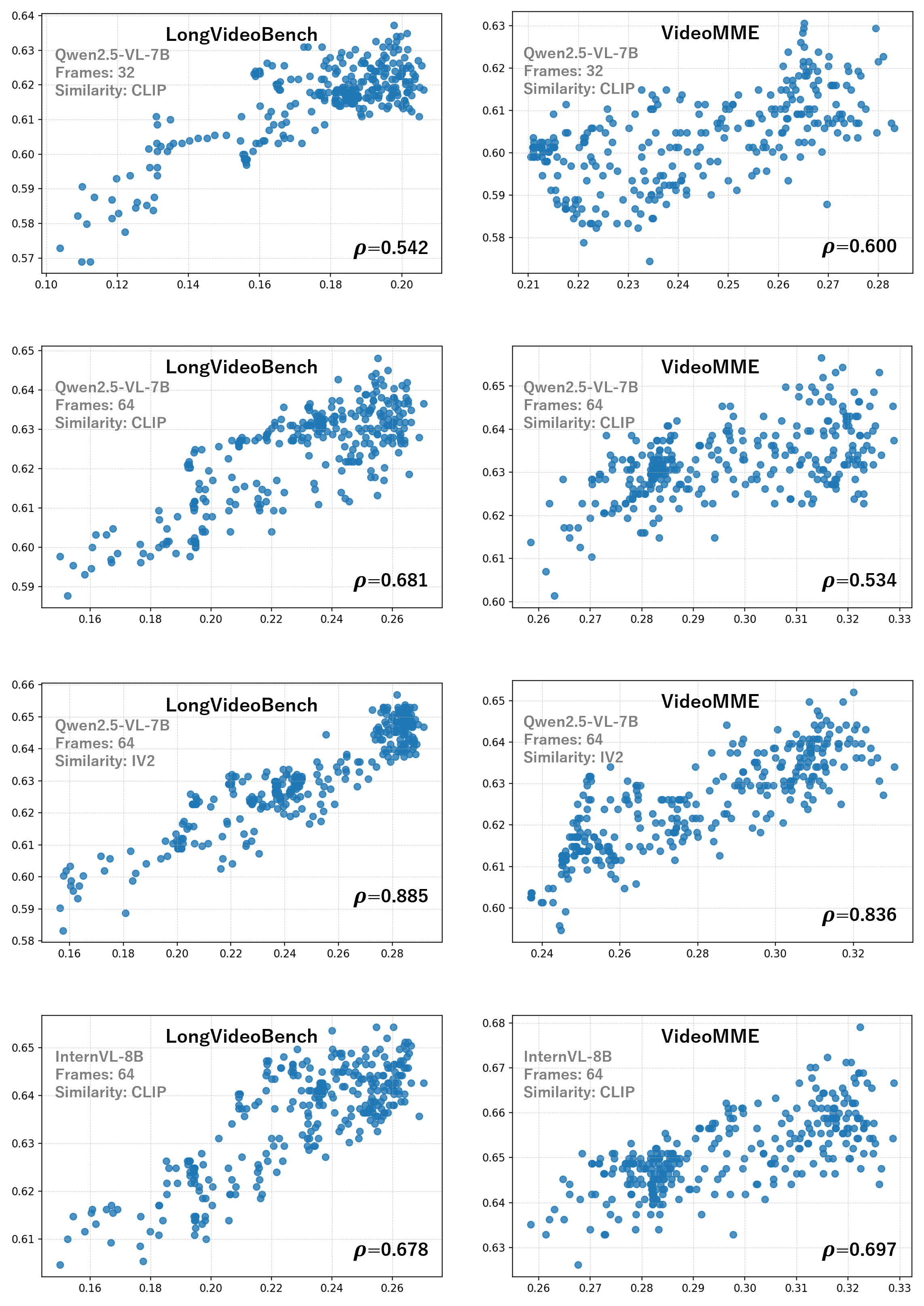}
    \vspace{-0.5em}
    \caption{Scatter plots of QA accuracy versus UKS. The x-axis represents UKSS values, and the y-axis represents QA accuracy. A clear positive correlation is observed, consistent with the correlation coefficients. While accuracy still varies within the *UKSS-optimal range* (UKSS within 0.01 of its maximum), the *accuracy-optimal range* (within 1\% of maximum accuracy) consistently corresponds to UKSS values close to their optimum, showing that UKSS can effectively narrow the search space and reduce evaluation cost.}
    \label{supp_fig1}
\end{figure*}

To provide an intuitive illustration of the correlation between QA accuracy and UKSS, we visualize the results as scatter plots. \figref{supp_fig1} shows the results under the four settings in Table~3 of the main paper, evaluated on both datasets. The x-axis represents UKSS values, and the y-axis represents QA accuracy. The plots reveal a clear positive correlation between accuracy and UKSS, consistent with the correlation coefficients reported in Table~3 of the main paper.

We further observe that when UKSS lies within 0.01 of its maximum value (referred to as the UKSS-optimal range), accuracy still exhibits noticeable variance. This indicates that UKSS alone cannot perfectly identify the sampling method or hyperparameters that yield the highest accuracy. However, when accuracy lies within 1\% of its maximum value (referred to as the accuracy-optimal range), UKSS is also found to be within or very close to its optimal range. This suggests that, when tuning sampling strategies with the goal of maximizing QA accuracy, UKSS can substantially narrow the search space and reduce the number of MLLM evaluations required, thereby lowering experimental cost.

\section{Additional Experiments on Factors Affecting QA Accuracy}
In addition to the experiments in Section~3.3 conducted on VideoMME with InternVL3-8B, we further performed experiments on LongVideoBench and with Qwen2.5-VL-7B. \tabref{supp_tab1} and \tabref{supp_tab2} report the effects of Key Frame Rate (KFR) and Scene Hit Rate (SHR) on QA accuracy when using InternVL3-8B and Qwen2.5-VL-7B, respectively. \tabref{supp_tab3} and \tabref{supp_tab4} show the effects of key frame distribution on QA accuracy under the same settings.

The observations from these results are almost entirely consistent with those in Section~3.3, with one minor exception: in \tabref{supp_tab2}, when controlling SHR on VideoMME, accuracy peaks at KFR = 60\% before slightly decreasing. However, since the decrease is marginal and such high KFR values rarely occur in practice, this phenomenon does not affect the design of the UKSS metric.

\section{Results by Single-Scene and Multi-Scene Samples}
\tabref{supp_tab7} reports the results separately for samples annotated with a single relevant scene and those annotated with multiple relevant scenes. We observe that our method achieves the best or second-best performance on most metrics, yielding the strongest overall QA and key frame sampling performance on both single-scene and multi-scene samples. However, our method underperforms ITS in UKSS on multi-scene samples of VideoMME, and its QA performance shows no clear advantage. Moreover, compared with the results reported in Section 5.3, the correlation between UKSS and QA accuracy is noticeably weaker on both single-scene and multi-scene samples, highlighting the limitations of UKSS and indicating that substantial room for improvement remains, particularly for multi-scene cases.

\begin{table*}[t]
    \caption{QA accuracy on \textbf{VideoMME}$_{\rm kfs}$ (left) and \textbf{LongVideoBench}$_{\rm kfs}$ (right) with different Key Frame Rates and Scene Hit Rates, where the number of frames per scene is proportional to scene duration. \textbf{MLLM: InternVL3-8B.}}
    \label{supp_tab1}
    \setlength{\tabcolsep}{1mm}
    \centering
    \begin{tabular}{c|c|cccccc}
        \toprule
        \multicolumn{2}{c|}{} & \multicolumn{6}{c}{Key Frame Rate} \\
        \cmidrule(lr){3-8}
        \multicolumn{2}{c|}{} & 0\% & 20\% & 40\% & 60\% & 80\% & 100\% \\
        \midrule
        \multirow{6}{*}{\rotatebox[origin=c]{90}{Scene Hit Rate}} & 0\% & 53.8 / 52.4 & - & - & - & - & - \\
        & 20\% & - & 64.6 / 64.6 & 66.7 / 66.6 & 67.3 / 67.8 & 66.8 / 69.1 & 67.3 / 71.2 \\
        & 40\% & - & 65.0 / 65.0 & 66.6 / 66.8 & 67.3 / 68.3 & 67.5 / 70.0 & 68.7 / 72.0 \\
        & 60\% & - & 65.3 / 65.3 & 67.6 / 67.4 & 67.8 / 68.9 & 68.2 / 70.5 & 68.8 / 73.3 \\
        & 80\% & - & 65.5 / 66.0 & 67.8 / 68.3 & 68.9 / 70.2 & 69.9 / 71.6 & 71.3 / 75.7 \\
        & 100\% & - & 66.9 / 66.9 & 68.7 / 69.1 & 70.0 / 70.9 & 71.2 / 72.9 & 73.2 / 77.8 \\
        \bottomrule
    \end{tabular}
\end{table*}
\begin{table*}[t]
    \caption{QA accuracy on \textbf{VideoMME}$_{\rm kfs}$ (left) and \textbf{LongVideoBench}$_{\rm kfs}$ (right) with different Key Frame Rates and Scene Hit Rates, where the number of frames per scene is proportional to scene duration. \textbf{MLLM: Qwen2.5-VL-7B.}}
    \label{supp_tab2}
    \setlength{\tabcolsep}{1mm}
    \centering
    \begin{tabular}{c|c|cccccc}
        \toprule
        \multicolumn{2}{c|}{} & \multicolumn{6}{c}{Key Frame Rate} \\
        \cmidrule(lr){3-8}
        \multicolumn{2}{c|}{} & 0\% & 20\% & 40\% & 60\% & 80\% & 100\% \\
        \midrule
        \multirow{6}{*}{\rotatebox[origin=c]{90}{Scene Hit Rate}} & 0\% & 49.7 / 51.2 & - & - & - & - & - \\
        & 20\% & - & 64.6 / 65.4 & 66.7 / 68.6 & 69.3 / 70.3 & 67.6 / 71.6 & 67.3 / 74.6 \\
        & 40\% & - & 65.3 / 65.7 & 67.5 / 69.3 & 70.5 / 71.2 & 68.7 / 72.1 & 68.8 / 75.3 \\
        & 60\% & - & 65.8 / 66.4 & 68.0 / 70.0 & 71.6 / 71.9 & 70.0 / 73.3 & 69.7 / 76.7 \\
        & 80\% & - & 66.1 / 66.5 & 68.5 / 70.5 & 72.1 / 73.0 & 71.3 / 74.8 & 71.1 / 79.2 \\
        & 100\% & - & 66.2 / 67.4 & 69.3 / 71.6 & 73.0 / 74.8 & 73.0 / 76.5 & 72.5 / 81.6 \\
        \bottomrule
    \end{tabular}
\end{table*}
\begin{table*}[t]
    \caption{QA accuracy on \textbf{VideoMME}$_{\rm kfs}$ (left) and \textbf{LongVideoBench}$_{\rm kfs}$ (right) under different distributions of key frames across scenes controlled by the parameters $C$ and $\beta$. \textbf{MLLM: InternVL3-8B.}}
    \label{supp_tab3}
    \setlength{\tabcolsep}{1mm}
    \centering
    \begin{tabular}{c|c|cccccc}
        \toprule
        \multicolumn{2}{c|}{} & \multicolumn{6}{c}{$\beta$} \\
        \cmidrule(lr){3-8}
        \multicolumn{2}{c|}{} & 0 & 0.2 & 0.5 & 1 & 2 & 5 \\
        \midrule
        \multirow{6}{*}{$C$} & 0.05 & 68.2 / 73.3 & 68.4 / 73.3 & 68.6 / 74.1 & 68.0 / 73.9 & 68.4 / 73.9 & 68.6 / 73.4 \\
        & 0.2 & 68.6 / 74.1 & 68.0 / 74.8 & 68.7 / 75.0 & 69.0 / 73.8 & 68.7 / 74.0 & 68.2 / 73.9 \\
        & 1 & 71.1 / 75.7 & 70.9 / 75.9 & 69.4 / 75.3 & 69.8 / 74.7 & 70.3 / 74.7 & 69.5 / 73.9 \\
        & 5 & 72.2 / 77.5 & 72.3 / 77.7 & 72.4 / 77.4 & 71.7 / 76.7 & 70.7 / 76.3 & 69.6 / 74.1 \\
        & 20 & 73.1 / 77.9 & 72.7 / 77.8 & 72.7 / 77.4 & 72.4 / 77.6 & 71.2 / 76.1 & 70.7 / 74.9 \\
        \bottomrule
    \end{tabular}
\end{table*}
\begin{table*}[t]
    \caption{QA accuracy on \textbf{VideoMME}$_{\rm kfs}$ (left) and \textbf{LongVideoBench}$_{\rm kfs}$ (right) under different distributions of key frames across scenes controlled by the parameters $C$ and $\beta$. \textbf{MLLM: Qwen2.5-VL-7B.}}
    \label{supp_tab4}
    \setlength{\tabcolsep}{1mm}
    \centering
    \begin{tabular}{c|c|cccccc}
        \toprule
        \multicolumn{2}{c|}{} & \multicolumn{6}{c}{$\beta$} \\
        \cmidrule(lr){3-8}
        \multicolumn{2}{c|}{} & 0 & 0.2 & 0.5 & 1 & 2 & 5 \\
        \midrule
        \multirow{6}{*}{$C$} & 0.05 & 68.1 / 76.1 & 68.0 / 75.9 & 68.1 / 77.0 & 67.9 / 77.4 & 68.2 / 77.8 & 68.0 / 77.6 \\
        & 0.2 & 68.6 / 77.4 & 68.0 / 77.5 & 68.1 / 77.6 & 68.7 / 77.0 & 68.0 / 77.6 & 67.5 / 77.7 \\
        & 1 & 69.8 / 79.7 & 69.8 / 79.7 & 68.4 / 79.2 & 68.8 / 79.2 & 68.7 / 79.1 & 68.6 / 78.4 \\
        & 5 & 72.1 / 81.0 & 72.5 / 80.8 & 70.9 / 80.9 & 72.2 / 79.8 & 70.8 / 79.8 & 69.3 / 78.4 \\
        & 20 & 72.2 / 81.2 & 72.5 / 81.6 & 72.0 / 81.2 & 72.4 / 81.0 & 71.8 / 79.8 & 70.3 / 78.7 \\
        \bottomrule
    \end{tabular}
\end{table*}
\begin{table*}[t]
    \caption{Experimental results on VideoMME and LongVideoBench (LVB). 
QA accuracy (acc.) and UKSS are reported separately for the single-scene and multi-scene samples of the KFS-Bench subsets.}
    \label{supp_tab7}
    \setlength{\tabcolsep}{1mm}
    \centering
    \begin{tabular}{ccccccccccc}
        \toprule
        \multirow{3}{*}{MLLM} & \multirow{3}{*}{Frames} & \multirow{3}{*}{\makecell{Sampling \\ Method}} & \multicolumn{4}{c}{\textbf{VideoMME}} & \multicolumn{4}{c}{\textbf{LVB}} \\
        \cmidrule(lr){4-7}
        \cmidrule(lr){8-11}
        & & & \multicolumn{2}{c}{\textit{Single}} & \multicolumn{2}{c}{\textit{Multi}} & \multicolumn{2}{c}{\textit{Single}} & \multicolumn{2}{c}{\textit{Multi}} \\
        \cmidrule(lr){4-5}
        \cmidrule(lr){6-7}
        \cmidrule(lr){8-9}
        \cmidrule(lr){10-11}
        & & & Acc. & UKSS & Acc. & UKSS & Acc. & UKSS & Acc. & UKSS \\
        \midrule
        \multirow{5}{*}{\makecell{Qwen2.5-\\VL-7B}} & \multirow{5}{*}{32} & Uniform & 63.7 & 0.198 & 46.0 & 0.211 & 59.5 & 0.083 & 51.2 & 0.161 \\
        & & K-means & 64.7 & 0.208 & 46.0 & 0.202 & 62.0 & 0.094 & \textbf{54.2} & 0.173 \\
        & & AKS & \underline{66.4} & 0.241 & 47.8 & \textbf{0.263} & 65.2 & 0.185 & 51.2 & 0.195 \\
        & & ITS & 64.5 & \underline{0.258} & \textbf{50.4} & \underline{0.256} & \underline{65.6} & \textbf{0.192} & 51.8 & \textbf{0.237} \\
        & & ASCS & \textbf{68.1} & \textbf{0.269} & \underline{49.3} & 0.255 & \textbf{67.7} & \underline{0.189} & \underline{52.4} & \underline{0.228} \\
        \midrule
        \multirow{5}{*}{\makecell{Qwen2.5-\\VL-7B}} & \multirow{5}{*}{64} & Uniform & 67.4 & 0.265 & 53.3 & 0.301 & 62.0 & 0.123 & 54.2 & 0.224 \\
        & & K-means & 67.1 & 0.278 & \underline{54.0} & 0.292 & 62.0 & 0.138 & 53.0 & 0.230 \\
        & & AKS & 69.2 & 0.278 & 51.1 & \underline{0.322} & \textbf{67.3} & 0.230 & 52.1 & 0.249 \\
        & & ITS & \underline{69.4} & \underline{0.304} & \textbf{55.1} & \textbf{0.334} & 66.5 & \underline{0.244} & \underline{54.5} & \underline{0.293} \\
        & & ASCS & \textbf{69.7} & \textbf{0.317} & 52.2 & 0.318 & \underline{67.1} & \textbf{0.257} & \textbf{56.1} & \textbf{0.308} \\
        \midrule
        \multirow{5}{*}{\makecell{Qwen2.5-\\VL-32B}} & \multirow{5}{*}{64} & Uniform & 69.5 & 0.265 & 55.1 & 0.301 & 61.4 & 0.123 & 57.0 & 0.224 \\
        & & K-means & 70.2 & 0.278 & \textbf{58.4} & 0.292 & 63.1 & 0.138 & \underline{57.6} & 0.230 \\
        & & AKS & \underline{71.3} & 0.278 & 54.4 & 0.322 & \underline{66.6} & 0.230 & 54.5 & 0.249 \\
        & & ITS & \textbf{71.5} & \underline{0.304} & 56.6 & \textbf{0.334} & 66.1 & \underline{0.244} & 56.7 & \underline{0.293} \\
        & & ASCS & 71.2 & \textbf{0.306} & \underline{58.0} & \underline{0.328} & \textbf{68.0} & \textbf{0.252} & \textbf{58.5} & \textbf{0.325} \\
        \midrule
        \multirow{5}{*}{\makecell{Intern\\VL3-8B}} & \multirow{5}{*}{64} & Uniform & 68.7 & 0.265 & 54.4 & 0.301 & 62.4 & 0.123 & 55.5 & 0.224 \\
        & & K-means & 69.1 & 0.278 & 55.8 & 0.292 & 64.4 & 0.138 & \underline{56.7} & 0.230 \\
        & & AKS & 69.9 & 0.278 & 56.2 & 0.322 & 67.4 & 0.230 & 54.2 & 0.249 \\
        & & ITS & \underline{71.2} & \textbf{0.304} & \underline{56.9} & \textbf{0.334} & \textbf{67.8} & \underline{0.244} & \underline{56.7} & \underline{0.293} \\
        & & ASCS & \textbf{71.3} & \underline{0.304} & \textbf{59.5} & \underline{0.329} & \underline{67.7} & \textbf{0.253} & \textbf{58.2} & \textbf{0.317} \\
        \bottomrule
    \end{tabular}
\end{table*}